\documentclass{article} %
\PassOptionsToPackage{table}{xcolor} %
\usepackage{iclr2027_conference,times}

\usepackage{amsmath,amsfonts,bm}

\def\eqref#1{equation~\ref{#1}}

\def\1{\bm{1}}

\DeclareMathAlphabet{\mathsfit}{\encodingdefault}{\sfdefault}{m}{sl}
\SetMathAlphabet{\mathsfit}{bold}{\encodingdefault}{\sfdefault}{bx}{n}

\usepackage[utf8]{inputenc} %
\usepackage[T1]{fontenc}    %
\usepackage{hyperref}       %
\usepackage{url}            %
\usepackage{booktabs}       %
\usepackage{amsfonts}       %
\usepackage{nicefrac}       %
\usepackage{microtype}      %
\usepackage[table]{xcolor}%

\usepackage{graphicx}
\usepackage{amsmath}
\usepackage{amssymb}

\usepackage{listings}
\usepackage{mwe} %
\usepackage{makecell}
\usepackage{color, colortbl}
\usepackage[normalem]{ulem} %
\usepackage{algorithm}
\usepackage{algorithmic}

\usepackage{wrapfig}
\usepackage{caption}
\usepackage{pifont} %
\usepackage{multirow}
\usepackage{booktabs}
\usepackage{float}

\usepackage{chngpage}
\usepackage{wrapfig}
\usepackage{adjustbox}
\usepackage{comment}
\usepackage{color}

\usepackage{listings}

\usepackage{bm}
\usepackage[capitalize]{cleveref}

\usepackage[toc]{appendix}
\usepackage{etoc}
\usepackage{minitoc}
\usepackage{mathtools}

\usepackage{amsthm}

\theoremstyle{definition}

\theoremstyle{remark}

\usepackage{tikz}
\usetikzlibrary{arrows.meta, calc, decorations.pathreplacing, angles, quotes, patterns, positioning}
\usepackage{subcaption}

\usepackage{wrapfig}

\crefname{section}{Sec.}{Secs.}
\Crefname{section}{Section}{Sections}
\Crefname{table}{Table}{Tables}
\crefname{table}{Tab.}{Tabs.}
\etocsettocstyle{\section*{Appendix}}{}

\definecolor{Gray}{gray}{0.9}
\definecolor{ImportantColor}{rgb}{0.63, 0.79, 0.95}
\newcolumntype{g}{>{\columncolor{ImportantColor}}c}
\newcolumntype{?}{!{\vrule width 1pt}}

\def\imw#1#2{\includegraphics[width=#2\linewidth]{#1.png}}

\newcommand{\tb}[3]{\setlength{\tabcolsep}{#2mm}\begin{tabular}{#1}#3\end{tabular}}

\graphicspath{
{figs/},
}

\newcommand*{\belowrulesepcolor}[1]{%
  \noalign{%
    \kern-\belowrulesep 
    \begingroup 
      \color{#1}%
      \hrule height\belowrulesep 
    \endgroup 
  }%
} 
\newcommand*{\aboverulesepcolor}[1]{%
  \noalign{%
    \begingroup 
      \color{#1}%
      \hrule height\aboverulesep 
    \endgroup 
    \kern-\aboverulesep 
  }%
}

\definecolor{codegreen}{rgb}{0,0.6,0}
\definecolor{codegray}{rgb}{0.5,0.5,0.5}
\definecolor{codepurple}{rgb}{0.58,0,0.82}
\definecolor{backcolour}{rgb}{0.95,0.95,0.92}

\lstdefinestyle{mystyle}{
    backgroundcolor=\color{backcolour},   
    commentstyle=\color{codegreen},
    keywordstyle=\color{magenta},
    numberstyle=\tiny\color{codegray},
    stringstyle=\color{codepurple},
    basicstyle=\ttfamily\footnotesize,
    breakatwhitespace=false,         
    breaklines=true,                 
    captionpos=b,                    
    keepspaces=true,                 
    numbers=left,                    
    numbersep=5pt,                  
    showspaces=false,                
    showstringspaces=false,
    showtabs=false,                  
    tabsize=2
}

\newcommand{\benchmark}{CRONOS}
\newcommand{\benchmarkfull}{\underline{C}ontinual \underline{R}obotic \underline{O}perations in \underline{NO}n-episodic \underline{S}ettings}

\title{\benchmark{}: Benchmarking Non-Episodic Reinforcement Learning for Robotic Manipulation}

\author{%
  Po-Yi Wu\footnotemark[1] \quad
  Djengo Cyun-Jyun Fang\footnotemark[1] \quad
  Dian Cheng\footnotemark[1] \quad
  Tsung-Wei Ke \\
  Department of Computer Science and Information Engineering\\
  National Taiwan University\\
  Taipei, Taiwan \\
  \texttt{\{B11902127, B09501048, B12902058, twke\}@csie.ntu.edu.tw}
}

\iclrfinalcopy

\begin{document}

\maketitle
{\renewcommand{\thefootnote}{\fnsymbol{footnote}}\footnotetext[1]{Equal contribution.}}
\lhead{Preprint} 

\begin{figure*}[ht!]
    \centering
    \includegraphics[width=\textwidth]{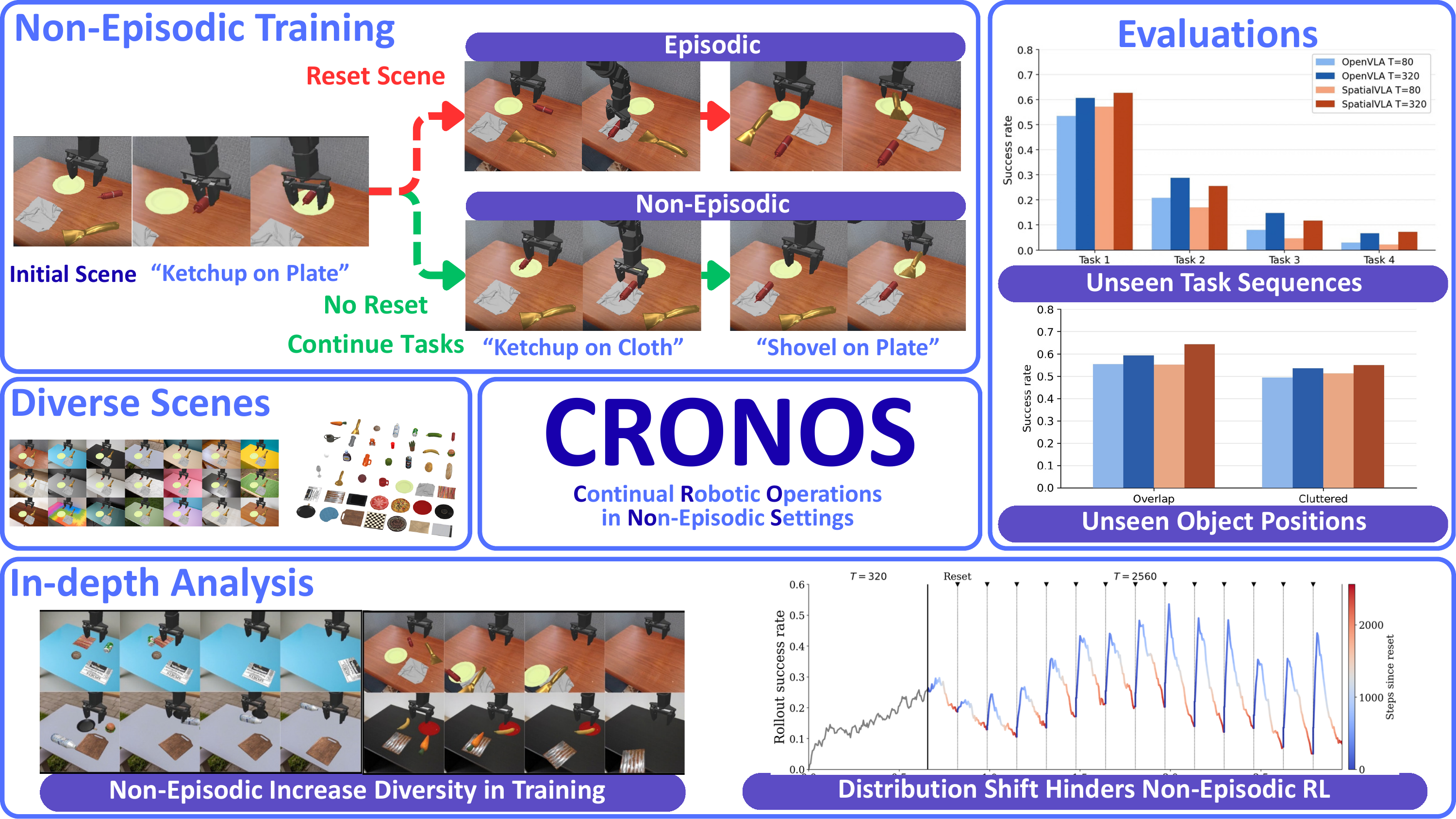}
    \caption{\small
    \textbf{Overview of non-episodic training in \benchmark{}.} We introduce \benchmark{}, a benchmark for RL fine-tuning of VLAs under infrequent scene resets. Tasks share a scene (top left). Episodic training resets the scene before every task, whereas non-episodic training continues each task from where the previous one ended. \benchmark{} covers diverse scenes and objects (middle left). Without frequent resets, object positions become increasingly dispersed, and rollout success decays after each scene reset (bottom). Yet policies fine-tuned non-episodically generalize better to unseen object positions and perform better on unseen task sequences (right).}
    \label{fig:teaser}
\end{figure*}

\begin{abstract}
Reinforcement learning (RL) is promising for adapting robot policies to unstructured real-world environments. However, standard RL pipelines rely on episodic training with frequent scene resets, which is impractical for real-world deployment due to the need for substantial human intervention. While prior work has studied non-episodic RL in relatively small policies trained from scratch, its behavior when fine-tuning modern large-scale pre-trained vision-language-action (VLA) models remains underexplored. In this work, we introduce \benchmark{}, a simulation benchmark for RL fine-tuning of VLAs under non-episodic scene resets. \benchmark{} leverages high-fidelity physics simulators and adopts shared-scene multi-task settings. Using \benchmark{}, we show that naively fine-tuning pre-trained policies fails in non-episodic settings due to shifts in object configurations and end-effector poses; however, these challenges can be mitigated through targeted reset allocation and by addressing biases in pre-trained models. Moreover, we find that non-episodic training enhances long-horizon manipulation and improves generalization to previously unseen object configurations and task sequences.
\end{abstract}

\section{Introduction}
\label{introduction}

Learning to adapt is key to deploying robots in continually changing worlds.  Recently, behavior cloning has emerged as a dominant learning approach in robotic manipulation.  It trains state-of-the-art visuomotor policies, often termed vision-language-action models (VLAs; \citealp{kim2024openvla,qu2025spatialvla,bjorck2025gr00t}), on large-scale expert demonstrations~\citep{o2024open,bu2025agibot}, enabling robots to perform complex manipulation tasks~\citep{black2024pi0,zheng2026x}.  However, VLAs are not robust to unseen conditions during training: even a slight change in lighting, object appearance or scene layout can cause a VLA's failure~\citep{barreiros2025behavior}.  Reinforcement learning (RL) offers a compelling solution.  RL adapts policies by trial and error, eliminating the persistent need for collecting demonstrations to cover new failures.  Recent studies show that RL fine-tuning substantially improves a VLA's generalization and robustness to novel scenarios~\citep{intelligence2025pi,zhang2026hy}.

Despite its generality, standard RL pipelines rely on an unrealistic assumption of \textit{episodic training}---scenes are frequently reset at each episode to prevent robots from being trapped in low-reward states during training, to enhance RL's training efficiency.  Because scene reset is free in simulation, this has been widely overlooked in prior work that studies RL fine-tuning in simulation~\citep{liu2025can,yu2026rlinf}.  In contrast, in real-world settings, scene reset requires human intervention and incurs substantial overhead.  This has prevented RL from being deployed at scale in real robotic systems.  To address the limitation, a wide range of non-episodic RL algorithms have been proposed, including manually scripting the reset robot policy~\citep{levine2016end,sharma2020emergent}, instrumenting the reset mechanism~\citep{zeng2020tossingbot,kalashnikov2018scalable,mtopt2021arxiv}, or tackling scene resets through a multi-task learning framework that either treats scene resets as additional tasks~\citep{eysenbach2018leave,zhu2020ingredients}, or decomposes a task into a set of subtasks, allowing the execution of each to reset the others~\citep{gupta2021reset}.  However, these approaches are proposed prior to the surge of VLAs, and confined to simpler settings of training policies from scratch.

We present an initial study to highlight challenges in RL fine-tuning VLAs under non-episodic scene reset, despite the prior success of episodic training.  We answer three questions: (1) how does RL fine-tuning VLAs perform out of the box, under non-episodic, multi-task settings? (2) what are the key factors impacting the performance? (3) what are the advantages of non-episodic training, aside from reducing the labor cost?  Experiments are primarily conducted in simulation benchmarks, as they are less time-consuming and offer better reproducibility than real-world benchmarks.

We introduce \benchmarkfull{} (\benchmark{}), a simulation benchmark carefully designed to facilitate systematic study on non-episodic scene reset RL.  \benchmark{} consists of a suite of 16 pick-and-place tasks, evenly distributed across 4 scenes.  It addresses multiple limitations of existing benchmarks.  First, \benchmark{} targets recent advances in VLAs.  It evaluates non-episodic RL fine-tuning of these pre-trained policies, rather than policies trained from scratch~\citep{gupta2021reset}.  Second, \benchmark{} adopts a realistic multi-task learning setup by sharing scenes across tasks, in contrast to prior benchmarks that isolate each task into a separate scene~\citep{yang2024robot}.  This compounds manipulation errors not only over the horizon, but also across tasks, as in real-world deployment where a robot keeps working in the same environment.  Third, \benchmark{} builds upon a high-fidelity physics simulator, Maniskill~\citep{tao2025maniskill3}, and its extension of manipulation tasks suites~\citep{li2024evaluating,liu2025can}.  In contrast, existing benchmarks rely on outdated simulators~\citep{balsells2023autonomous} with simplified physics and limited visual realism.  Both our multi-task learning setup and usage of modern simulators make non-episodic scene reset RL more challenging.  We present an overview comparison between common multi-task episodic RL and our shared-scene multi-task non-episodic scene reset RL in Figure~\ref{fig:teaser}.

We test two off-the-shelf VLAs: OpenVLA~\citep{kim2024openvla} and SpatialVLA~\citep{qu2025spatialvla}, on \benchmark{}.  We arrive at following findings: \textbf{(1) Non-episodic RL fine-tuning for pre-trained VLAs is ineffective.}  We show that distribution drifts in initial scene layout and initial robot configurations make non-episodic training inefficient, \textbf{(2) Targeted resets improves non-episodic training.}  We found that resetting strategies, that handle the distribution drifts, can improve non-episodic RL's efficiency, while others show no benefit.  \textbf{(3) Non-episodic training brings stronger out-of-domain generalization.}  While difficult to train, we found that policies fine-tuned with non-episodic RL are more robust to novel object configurations and long-horizon robotic manipulation, than their episodically fine-tuned counterparts.

\section{Preliminaries}
\label{sec:preliminary}

In this section, we present a general formulation of multi-task RL problems, and extend it to non-episodic scene reset multi-task RL (NSR-MTRL) problems without loss of generality.

\paragraph{Multi-task RL.}
Multi-task RL problems consider learning optimal policies for a set of tasks sampled from a task distribution $p_z$.  A task $z \sim p_z(\cdot)$ defines a Markov decision process (MDP) $\mathcal{M}_z = (\mathcal{S}, \mathcal{A}, P, R_z, \rho, \gamma)$, where $\mathcal{S}$ is the state space, $\mathcal{A}$ is the action space, $P$ denotes the transition dynamics of the environment, $R_z$ is the reward function of the task, $\rho$ is the initial state distribution, and $\gamma$ denotes the discount factor. Given policy $\pi$ conditioning on task $z$, the RL objective $J(\pi)$ is to maximize the expected return for every task:
\begin{align}
    J(\pi) = \mathbb{E}_{z \sim p_z(\cdot)} \mathbb{E}_{s_0 \sim \rho, a_h \sim \pi(\cdot | s_h; z), s_{h+1} \sim P(\cdot| s_h, a_h)}\left[ \sum_{h} \gamma^h R_z(s_h, a_h) \right].
    \label{eqn:rl_objective}
\end{align}

\paragraph{Non-episodic Scene Reset Multi-Task RL (NSR-MTRL).}
We extend this formulation to \textit{non-episodic scene reset} settings.  Evaluation remains episodic, with trajectories rolled out from the initial state for each task; whereas, training trajectories are rolled out continually across tasks with no or only occasional scene resets~\citep{sharma2022autonomous}.  Therefore, the evaluation objective is identical to Eqn.~\ref{eqn:rl_objective}, but the training objective differs.
To model occasional scene resets, we define $\varepsilon$ as the reset frequency and re-write the transition dynamics as: $\hat{P}(\cdot|s_h, a_h) = (1 - \varepsilon) P(\cdot|s_h, a_h) + \varepsilon \rho(\cdot)$, where the next state follows the environment dynamics $P$ with probability $1-\varepsilon$, and is instead sampled from the initial-state distribution $\rho$ with probability $\varepsilon$.
Let $z_h$ denote the task at time $h$, we follow ~\cite{gupta2021reset} and write the training objective of NSR-MTRL as:
\begin{align}
    \hat{J}(\pi) = \mathbb{E}_{s_0 \sim \rho} \mathbb{E}_{z_h \sim p_z(\cdot), a_h \sim \pi(\cdot | s_h; z_h), s_{h+1} \sim \hat{P}(\cdot| s_h, a_h)}\left[ \sum_{h} \gamma^h R_{z_h}(s_h, a_h) \right].
    \label{eqn:reset_free_rl_objective}
\end{align}

\section{Non-episodic scene reset trades interaction efficiency for reset efficiency}
\label{sec:benchmark}
To find out if RL fine-tunes VLAs effectively under non-episodic multi-task settings, we propose a challenging simulation benchmark and evaluate NSR-MTRL fine-tuning for VLAs on the benchmark.

\subsection{A Challenging Benchmark for NSR-MTRL}

We introduce \benchmark{}, a simulation benchmark designed to study non-episodic scene reset multi-task RL (NSR-MTRL).  It is built upon two established robotic manipulation simulation benchmarks, SimplerEnv~\citep{li2024evaluating} and its extension~\citep{liu2025can}. Both adopt a modern physics simulator, Maniskill3~\citep{tao2025maniskill3}, which produces high-fidelity physics simulation.

\textbf{Training}. Unlike previous benchmarks isolating each task into a separate scene, \benchmark{} adopts shares scenes across tasks, as shown in Figure~\ref{fig:teaser} (top-left).  This makes multi-task learning more challenging, because manipulation errors propagate to next tasks.  To enable this, \benchmark{} creates four scenes with different environment backgrounds, table meshes, and lighting conditions. Each scene consists of a different set of 2 objects and 2 receptacles, yielding 4 language-conditioned tasks of the form \textit{``put the [object] on the [receptacle]''} per scene, for a total of 16 tasks across all four scenes. Each task is rolled out for 80 interaction steps in the environment. In episodic training, we set the scene reset interval $T$ to 80 interaction steps. Upon scene reset, the object and receptacle are randomly placed in one of 16 fixed positions, immediately before the policy switches to rolling out the next task. In non-episodic training, we reset scene every $T = 320$ or $T = 2560$ steps, corresponding to a reset after every 4 or 32 tasks, respectively.

\textbf{Evaluation.} We adopt success rate, the fraction of trials that end in success, as the primary metric.  
We evaluate policies under two protocols, independent-task evaluation and sequential-task evaluation, each on both object and receptacle spatial configurations seen (In Domain) and unseen (Out of Domain) during training. For independent-task evaluation, the policy has up to 80 steps to complete a single task; the trial succeeds if it does so within that limit. For sequential-task evaluation, the policy is instructed to complete a sequence of tasks, each with an 80-step budget. A task not completed in time is counted as a failure, and the policy moves on to the next one. We evaluated sequential tasks performance in both training and unseen task sequences.

In most experiments, we adopt independent-task evaluation in In Domain object and receptacle spatial configurations, unless otherwise mentioned.

\textbf{Models.}
In this work, we study two off-the-shelf VLAs: OpenVLA~\citep{kim2024openvla} and SpatialVLA~\citep{qu2025spatialvla}.  The former is widely adopted in previous approaches to RL fine-tuning~\citep{zhang2024grape,liu2025can,yu2026rlinf}, and the latter shows strong performance in robotic manipulation.  Both models are trained with PPO algorithm~\citep{schulman2017proximal}.

\begin{figure}[H]
    \centering
    \begin{adjustbox}{width=0.9\linewidth}
    \tb{@{}c@{\hspace{3em}}c@{}}{0.1}{
    \imw{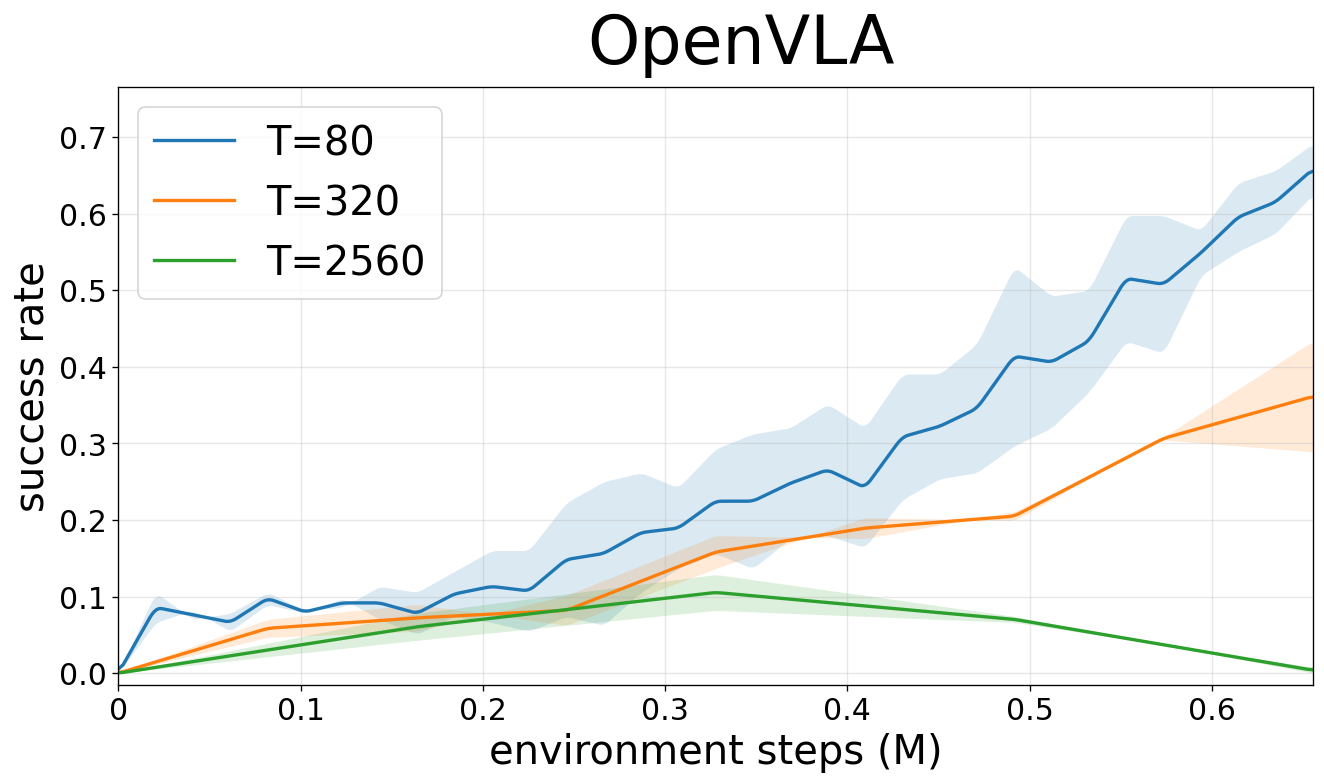}{0.45} &
    \imw{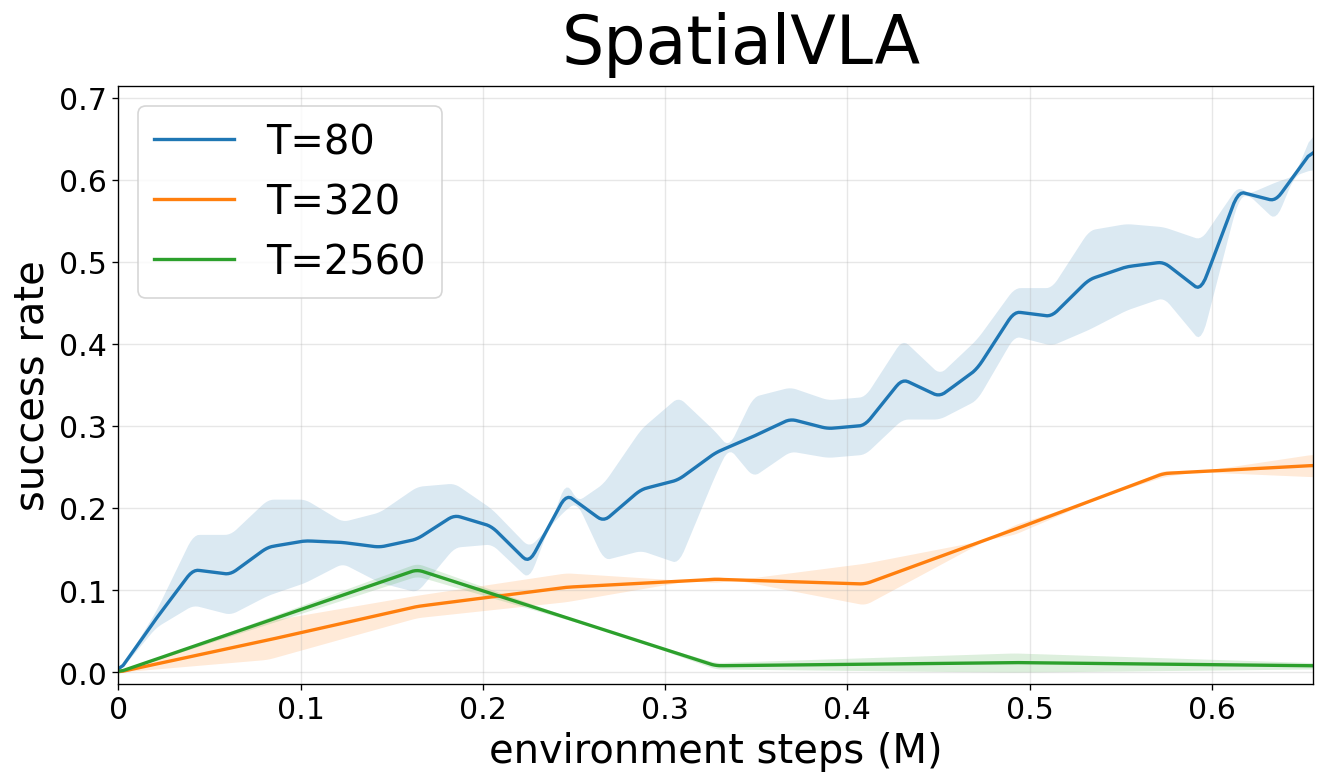}{0.45} \\
    }
    \end{adjustbox}
    \caption{\textbf{Pretrained VLAs struggle to learn under non-episodic scene resets.} When directly fine-tuned in the non-episodic multi-task setting, learning performance dropped dramatically as the scene reset interval $T$ increased from 80 to 320 interaction steps, and failed at $T = 2,560$ steps.}
    \label{fig:Q1}
\end{figure}

\subsection{Pretrained VLAs Struggle to Learn under Non-Episodic Scene Resets}
\label{sec:findings}

We summarize our preliminary study in Figure~\ref{fig:Q1}.  Our first conclusion is that directly fine-tuning VLAs under non-episodic multi-task settings is ineffective.  NSR-MTRL's learning efficiency substantially drops with decreasing frequency of scene reset: within training scenes, OpenVLA's success rate drops from $65\%$ to $37\%$ and $0\%$ with scene reset interval $T$ ranging from $80$ to $320$ and $2560$; likewise, SpatialVLA's success rate drops from $65\%$ to $25\%$ and $2\%$.  Both models show similar trends when tested on Out of Domain configurations.

While adapting VLAs efficiently, episodic RL requires frequent but laborious scene resets, which hinder episodic RL's application on real-world robotic systems.  This necessitates NSR-MTRL algorithms, which should enhance both the reset and interaction efficiency for fine-tuning VLAs.  Next, we conduct a series of experiments to study the problem in depth.  We aim to answer two questions: (1) Can different reset mechanisms impact NSR-MTRL performance? and (2) what are the strengths of NSR-MTRL fine-tuned policies, compared to their episodically trained counterpart?

\begin{figure}[H]
\centering
\begin{subfigure}[t]{0.325\textwidth}
    \centering
    \includegraphics[height=3.5cm]{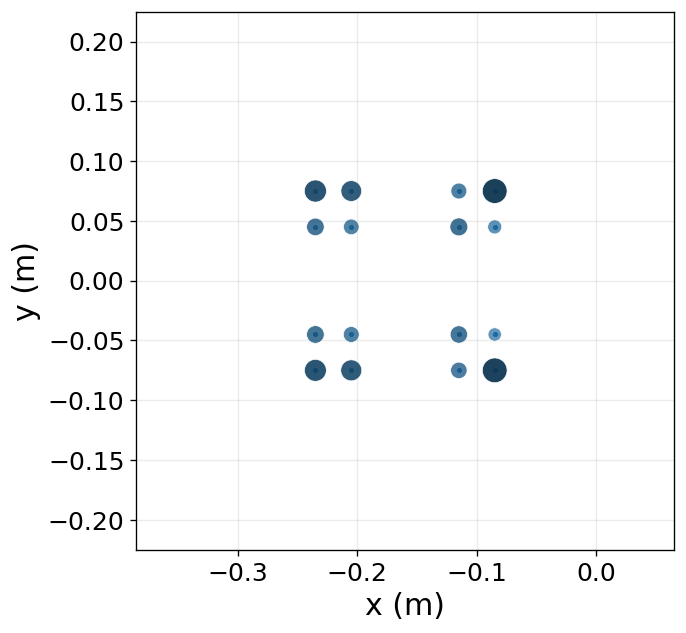}
    \caption{$T=80$}
    \label{fig:T80_obj_distribution}
\end{subfigure}
\hfill
\begin{subfigure}[t]{0.325\textwidth}
    \centering
    \includegraphics[height=3.5cm]{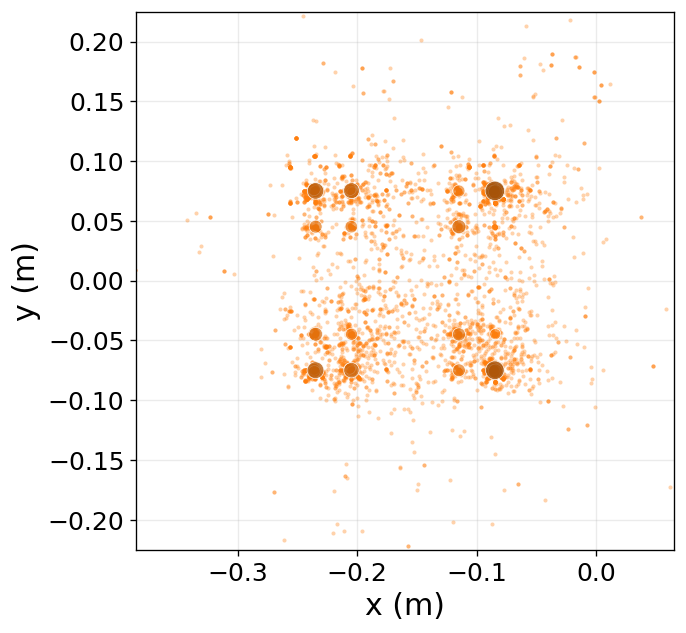}
    \caption{$T=320$}
    \label{fig:T320_obj_distribution}
\end{subfigure}
\hfill
\begin{subfigure}[t]{0.325\textwidth}
    \centering
    \includegraphics[height=3.5cm]{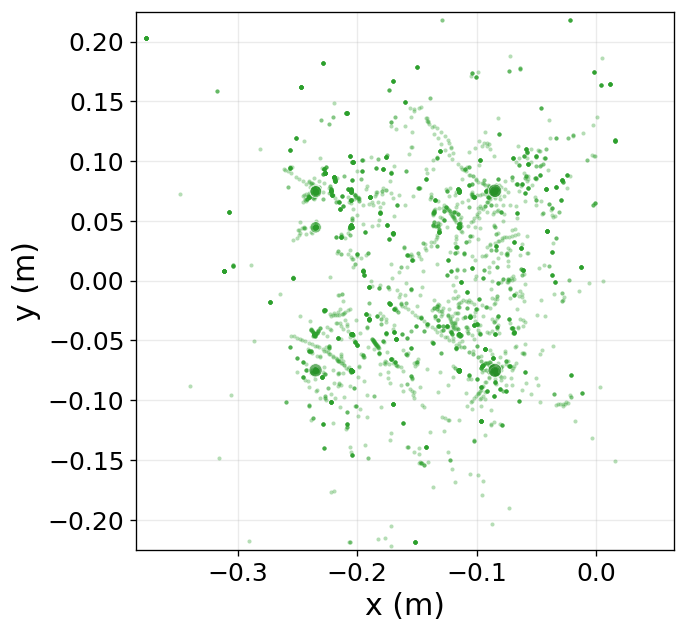}
    \caption{$T=2560$}
    \label{fig:T2560_obj_distribution}
\end{subfigure}
\caption{\textbf{Long scene reset intervals expose the policy to diverse object starting positions.} In non-episodic RL, each new task starts with the scene layout carried over from the previous task, causing object starting positions to become increasingly dispersed as the scene reset interval increases.}
\label{fig:obj_distribution}
\end{figure}

\begin{figure}[t]
    \centering
    \begin{minipage}[t]{0.48\linewidth}
        \centering
        \imw{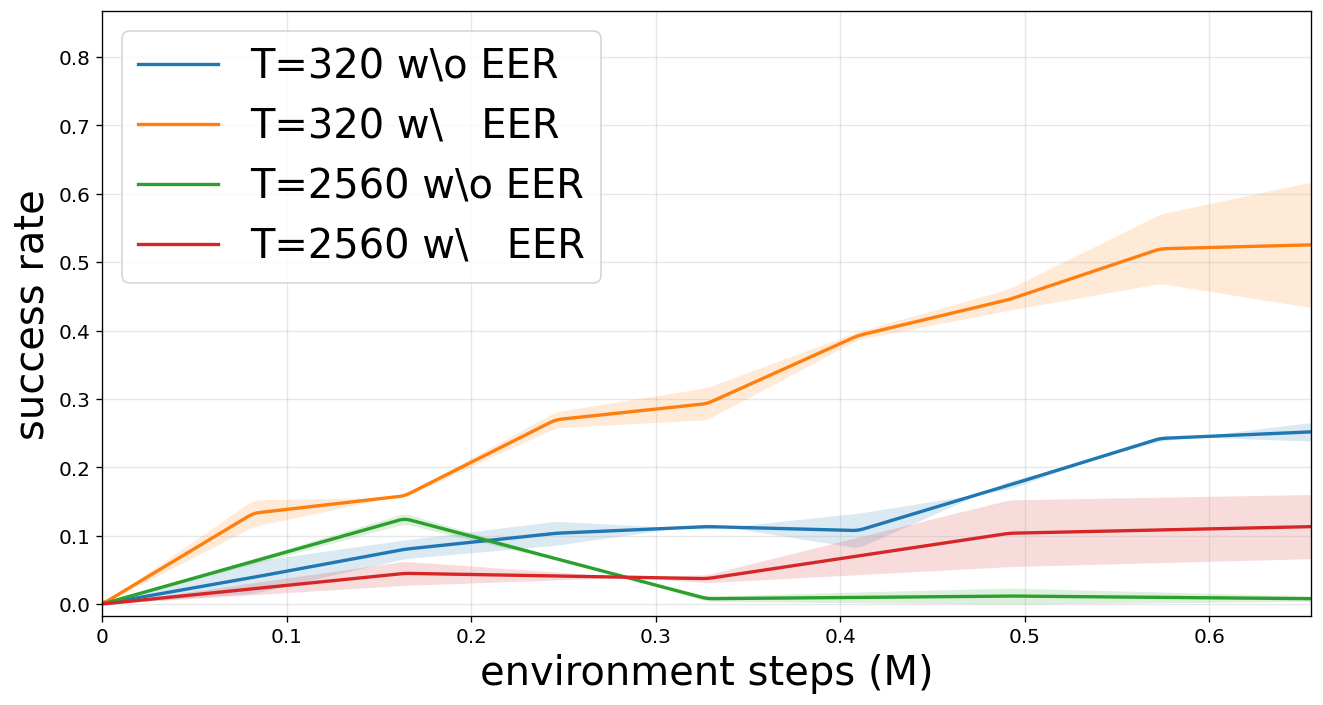}{1.0}
        \captionof{figure}{\textbf{End-effector reset mitigates pretraining bias.} EER automatically resets the end-effector to its initial position at each task switch, improving the learning efficiency of SpatialVLA under non-episodic scene-reset settings. }
        \label{fig:Q2}
    \end{minipage}
    \hfill
    \begin{minipage}[t]{0.48\linewidth}
        \centering
        \imw{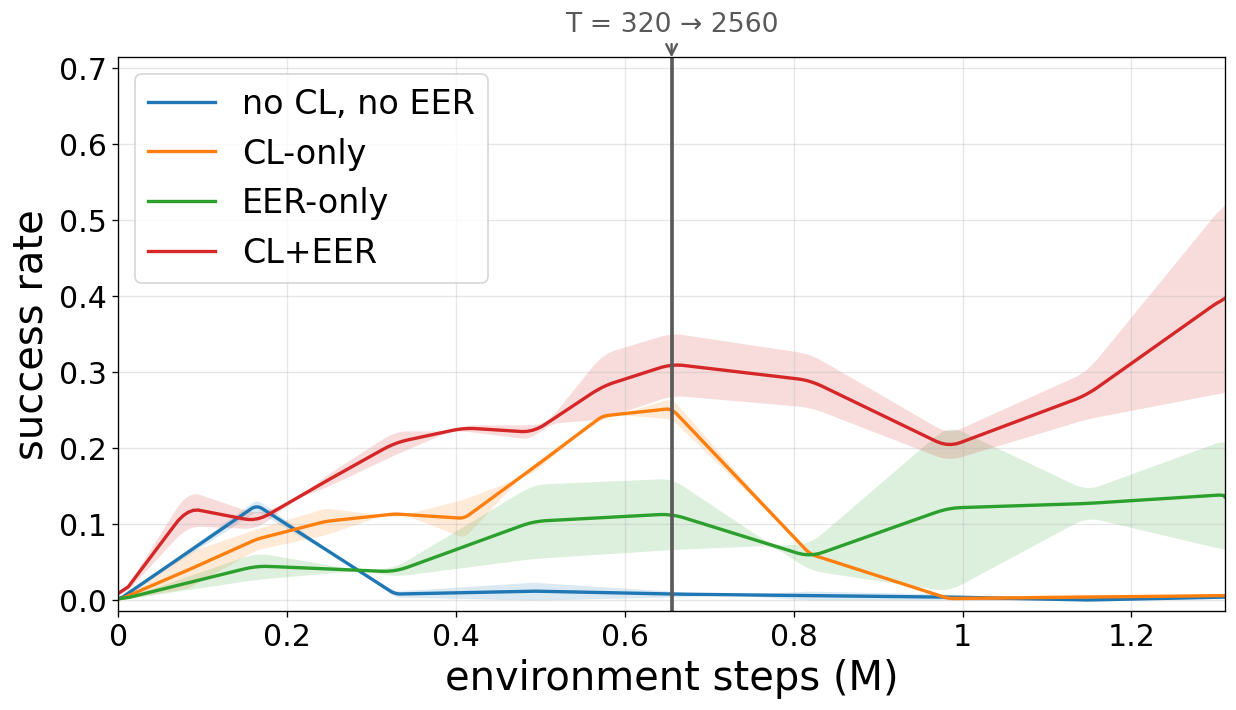}{1.0}
        \captionof{figure}{\textbf{Frequent scene resets during early training improve interaction efficiency.} With EER, training SpatialVLA with a reset interval of 320 steps followed by an increase to 2560 steps is the only strategy that exhibits an upward trend in task success rate.}
        \label{fig:Q3}
    \end{minipage}
\end{figure}

\section{Targeted Resets Enable Non-Episodic Scene Reset Fine-Tuning }
\label{experiments}

As shown in Figure~\ref{fig:Q1}, reducing scene reset frequency directly degrades interaction efficiency.  This raises the question of whether improving NSR-MTRL's reset mechanisms achieves better tradeoff between reset and interaction efficiency.  We investigate three approaches:  (1) resetting only the robot arm, which can be performed automatically without human intervention; (2) adjusting the scene reset frequency throughout training; and (3) resetting low-reward scenes to prevent the loss of learning signals from unrecoverable states.

\subsection{End-Effector Resets Reduce Initial-State Distribution Drift}
\label{subsec:bias}

In large-scale demonstration datasets, such as Open X-Embodiment~\citep{o2024open} and RH20T~\citep{fang2023rh20t} used for pre-training OpenVLA and SpatialVLA, demonstrations typically start from a fixed end-effector pose.  This yields a strong bias of the robot's initial pose distribution, that is later learned by VLAs.  In contrast, during non-episodic training, a new task starts with both the scene layout and the robot configuration continued from the end of the previous task, which drift as each task proceeds.  Motivated by this observation, we study whether the distribution shift in the robot's end-effector position impacts NSR-MTRL's fine-tuning efficiency.

To eliminate the distribution shift, we introduce episodic end-effector reset (EER): at each task transition, it returns the end-effector automatically to its canonical pose, while leaving the scene untouched.  Unlike scene reset, EER does not need any human intervention, and it can be performed using inverse kinematics or motion planning.

Surprisingly, we found that EER substantially enhance NSR-MTRL's efficiency.  As shown in Figure~\ref{fig:Q2}, EER improves SpatialVLA's success rate by 25 and 10 percentage points with scene reset interval $T=320$ and $T=2560$.  \textbf{Our results demonstrate that pre-trained VLAs are biased by the robot's initial pose, and reducing the distribution drift is key to NSR-MTRL's performance.} We therefore use EER in all subsequent experiments.

\subsection{Frequent Scene Resets Early Improve NSR-MTRL Efficiency}
\label{subsec:cl}
While EER reduces distribution drift in initial robot pose, it does not handle distribution drift in initial scene layout.  We next study how much drift in initial scene layout results from non-episodic training, and whether reducing the drift in the early training stage enhances NSR-MTRL's efficiency.

First, we analyze the drift in initial scene layout.  We record the positions of objects and receptacles at the start of each task during training.  For episodic training, each object or receptacle is placed at one of the 16 positions.  For non-episodic training, objects and receptacles are moved by the robot to random positions.  As shown in Figure~\ref{fig:obj_distribution}, the initial scene layout becomes more diverse with less frequent scene resets, and the trend coincides with the degrading efficiency of NSR-MTRL.

Next, we study the impact of reducing the distribution drift in initial scene layout.  We propose a curriculum-learning (CL) approach that performs scene resets frequently early in training, and reduces reset frequency as training proceeds.  Our idea is training policy to handle the distribution gap by gradually exposing it to novel distributions. We adopt a simple two-stage schedule and leave more advanced designs to future work.  We first train VLAs with $T=320$ for the first 600K steps and then increase the interval to $T=2560$.  We compare CL fine-tuned policies against their counterparts without CL fine-tuning. As shown in Figure~\ref{fig:Q3}, CL significantly enhances NSR-MTRL fine-tuning for VLAs.  Tested on both in-domain and out-of-domain configurations, SpatialVLA fine-tuned with CL achieves $40\%$ and $32\%$ success rate, compared to $14\%$ and $8\%$ by the baseline.  \textbf{Our results demonstrate that allocating scene resets more intelligently is key to NSR-MTRL's performance.}

\begin{figure*}[t]
    \centering
    \includegraphics[width=0.9\textwidth]{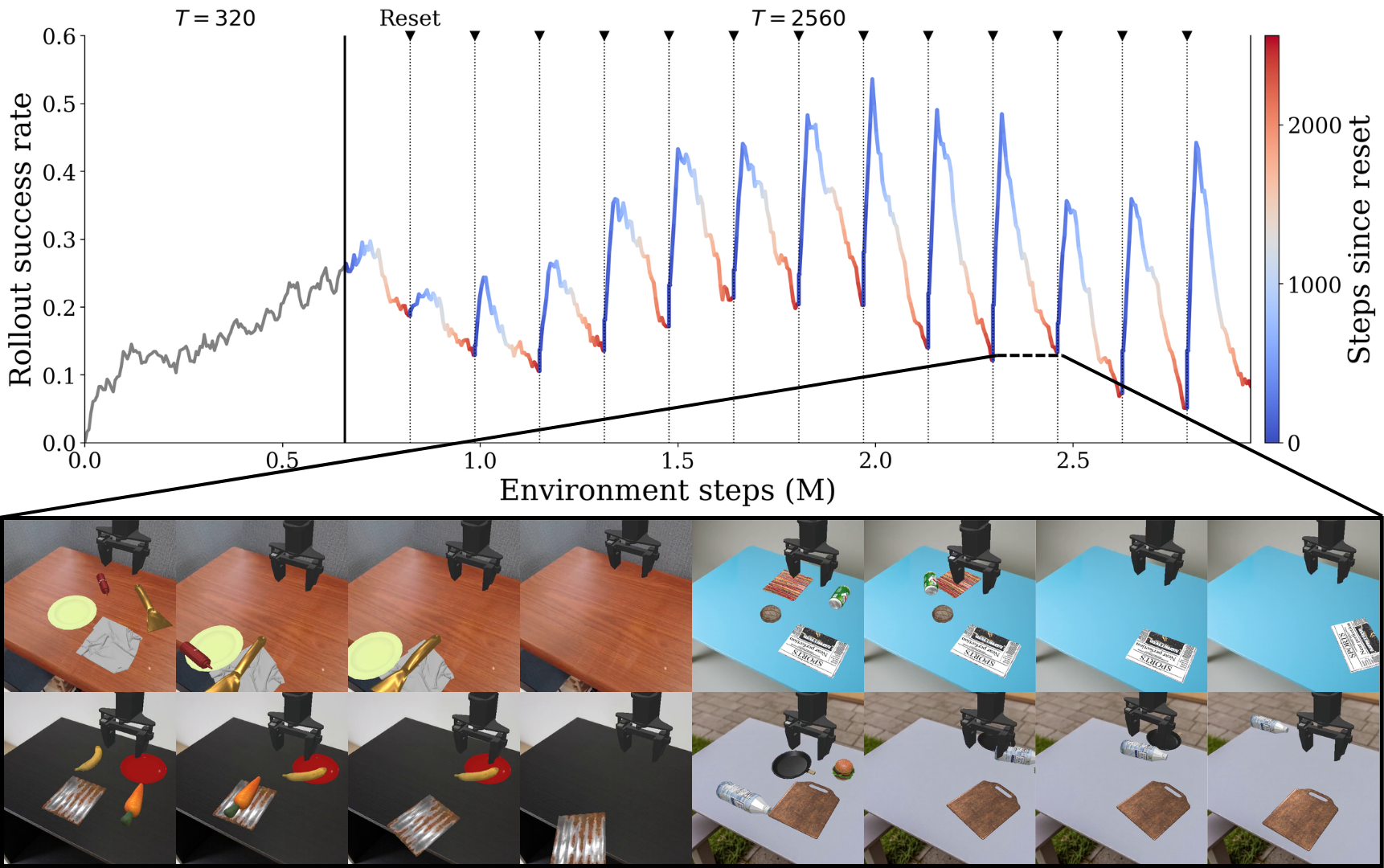}
    \caption{\textbf{Rollout success decays between scene resets.} Top. Rollout success rate of SpatialVLA during non-episodic training. The vertical line at step 600K marks the switch from $T=320$ to $T=2560$, and triangles mark scene resets. Bottom. Snapshots of the scene between two consecutive resets. Objects become increasingly dispersed and eventually leave the reachable workspace, leaving few successful rollouts to learn from.}
    \label{fig:rollout_success}
\end{figure*}

\subsection{Resetting Unrecoverable Scenes Prevent Loss of Learning Signals}
Even with EER and CL, learning under sparse scene resets remains challenging.  As shown in the top of Figure~\ref{fig:rollout_success}, the policy's success rate peaks right after every scene reset, then decays along time.  We visualize the scene evolving between two scene resets in the bottom of Figure~\ref{fig:rollout_success}.  We observe that object positions become increasingly dispersed, and eventually they fall out of the robot's reachable workspace.  Because the robot is stuck in unrecoverable states, the policy can never complete the task and therefore obtains zero reward from collected rollout.  Human intervention is needed in such cases.

\begin{figure}[t]
    \centering
    \begin{minipage}[t]{0.48\linewidth}
        \centering
        \begin{adjustbox}{width=\linewidth}
        \imw{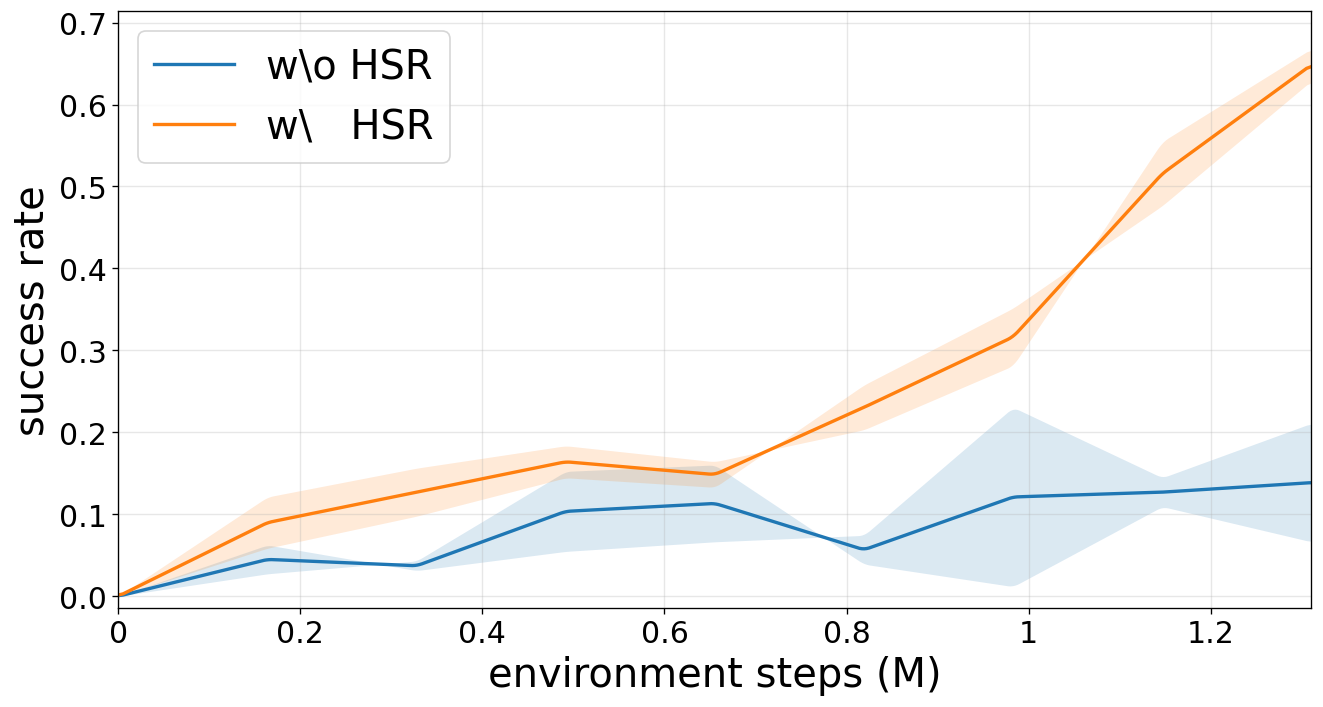}{1.0}
        \end{adjustbox}
        \caption{\textbf{Resetting unrecoverable scenes preserves learning signals.} By resetting all unrecoverable scenes at each task switch, HSR achieved a 64\% task success rate, compared with only 14\% success rate without HSR.}
        \label{fig:HSR}
    \end{minipage}
    \hfill
    \begin{minipage}[t]{0.48\linewidth}
        \centering
        \imw{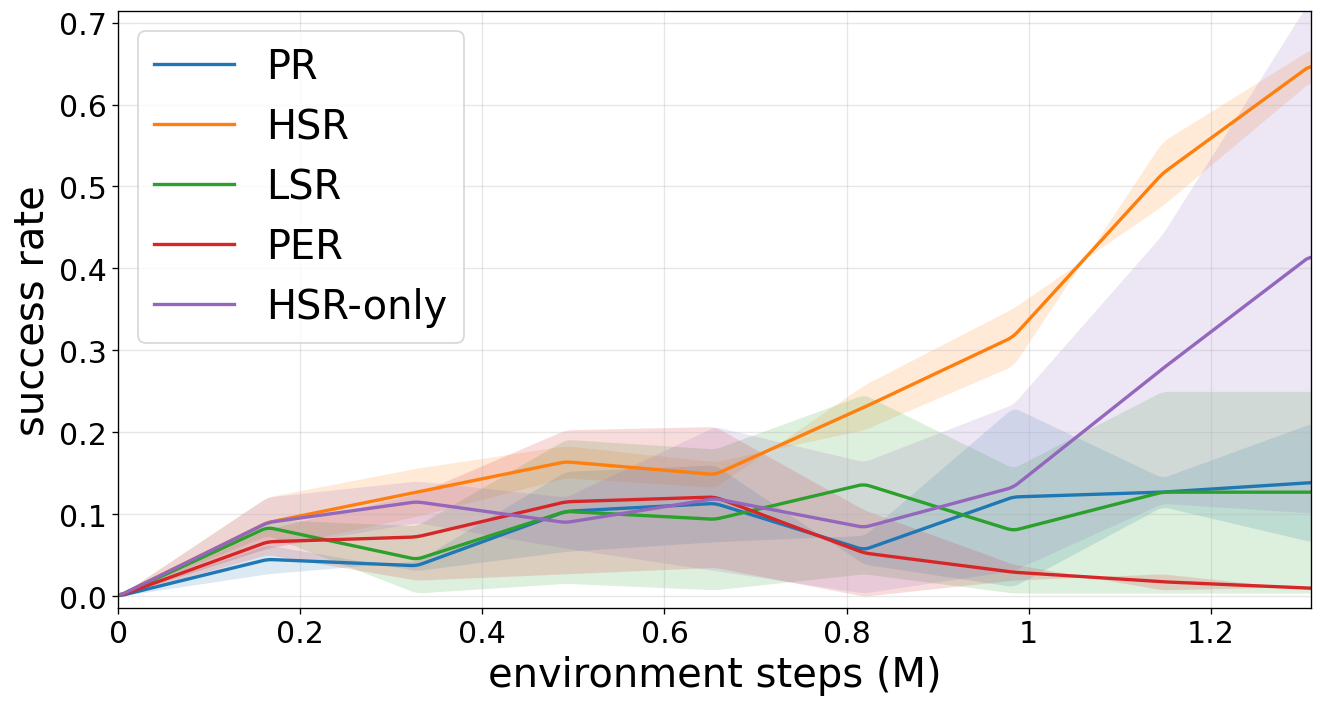}{1.0}
        \caption{\textbf{Forced reset of unrecoverable scenes outperforms learned reset strategies.} Only methods that forcibly reset all unrecoverable scenes (HSR and HSR-only) show a sustained increase in success rate.}
        \label{fig:Q4}
    \end{minipage}
\end{figure}

To resolve this, we first introduce a heuristic approach, Heuristic Scene Reset (HSR).  With scripted functions, it checks whether the current scene is unrecoverable at each task transition.  If any object falls off the table or lies outside the robot's workspace, HSR triggers a scene reset; otherwise, the policy continues from the current scene, proceeding to the next task.  We test SpatialVLA trained with HSR and scene reset interval $T=2560$.  As shown in Figure~\ref{fig:HSR}, HSR improves SpatialVLA's success rate from $14\%$ to $64\%$ on in-domain positions .  Motivated by these results, we explore other ways to introduce scene resets.

We study four alternatives to the heuristic approach: (1) Periodic Reset (PR), that resets the scene periodically and it is the vanilla approach adopted by NSR-MTRL, (2) HSR-only, that applies HSR without PR, (3) Learned Scene Reset (LSR; \citealp{eysenbach2018leave}), that additionally trains the policy to relocate the same object as the previous task to any valid region on the table.  By learning to reset objects to valid configurations, LSR can avoid the robot stuck in unrecoverable states, and (4) Perturbation Policy (PER; \citealp{zhu2020ingredients}), that expands LSR by relocating the same object as the previous task to any valid region on the table or to another receptacle.  Except HSR-only, all other methods also perform PR.  More details are provided in Appendix~\ref{app:reset_methods}.

\begin{figure}[H]
    \centering
    \begin{adjustbox}{width=0.9\linewidth}
    \tb{@{}c@{\hspace{3em}}c@{}}{0.1}{
    \imw{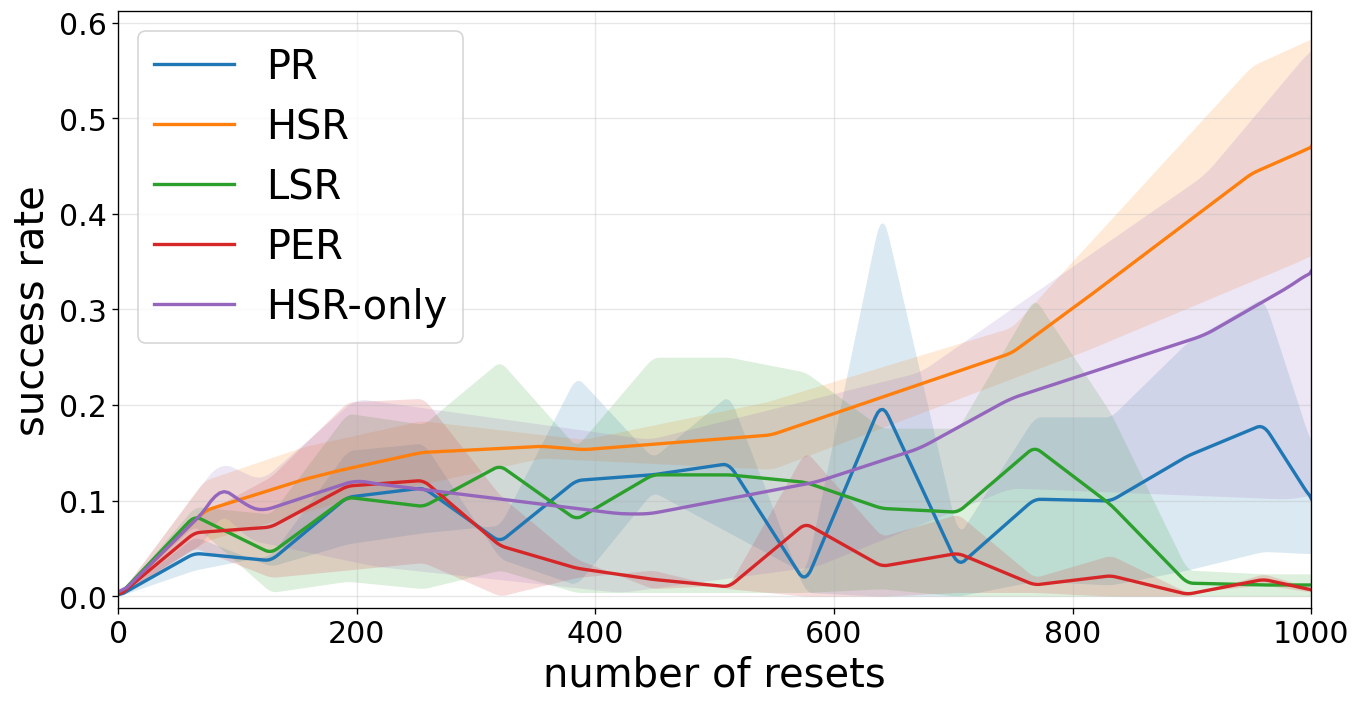}{0.45} &
    \imw{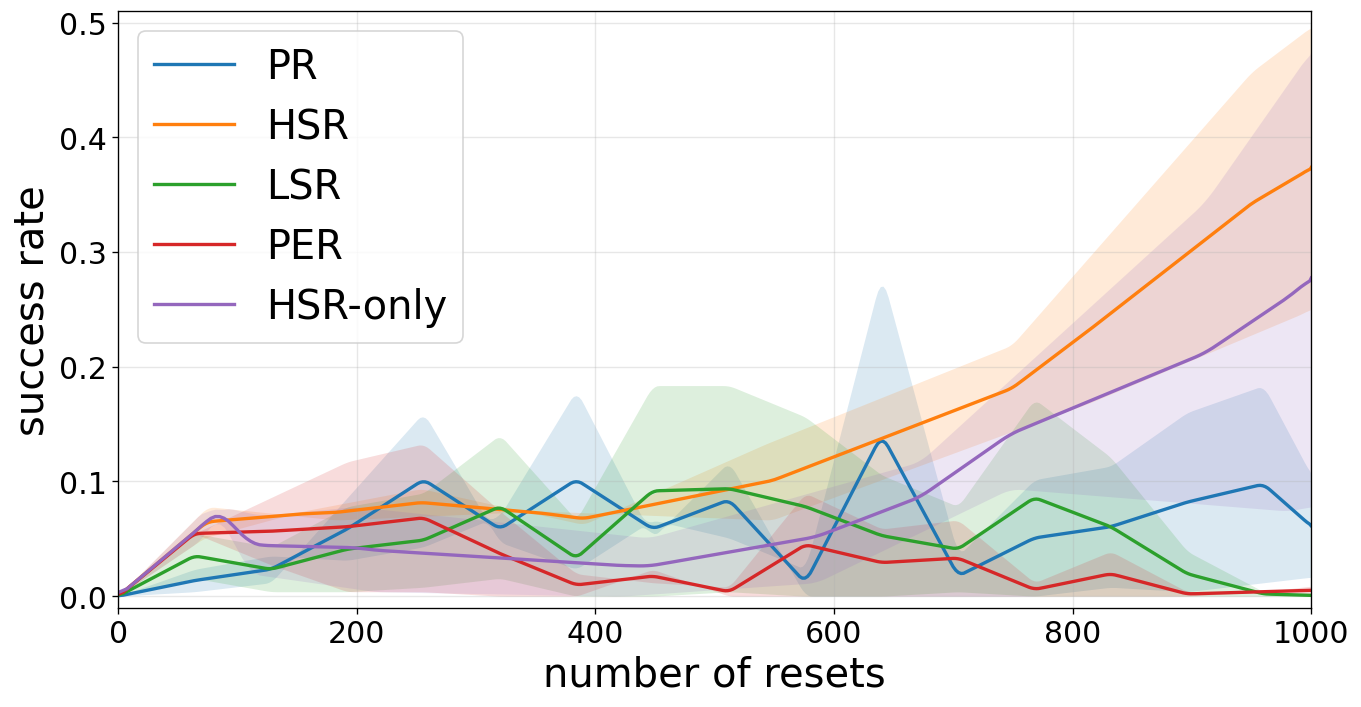}{0.45} \\
    
    }
    \end{adjustbox}
    \caption{\textbf{Forcibly resetting unrecoverable scenes is the most reset-efficient strategy.} As the number of scene resets increases, only methods that forcibly reset all unrecoverable scenes (HSR and HSR-only) continue to improve in performance.}
    \label{fig:Q3_reset}
\end{figure}

We conduct experiments with SpatialVLA.  Aside from their individual reset strategies, PR, HSR, LSR and PER also periodically reset the scene with interval $T=2560$.  As shown in Figure~\ref{fig:Q4}, only two approaches---HSR and HSR-only---demonstrate positive impact on NSR-MTRL; LSR performs as poorly as PR, and PER fails to adapt SpatialVLA to any task.  Our results are not surprising.  Both LSR and PER train policies to perform additional tasks, which are as inefficient to learn as the original pick-and-place tasks.  Moreover, the robot can still enter unrecoverable states when training these reset tasks.  In contrast, both HSR and HSR-only intervene to avoid unrecoverable states. One concern is that HSR may trigger more resets, trading reset efficiency for interaction efficiency. To examine this, we compare all methods under a fixed budget of 1000 resets. As shown in Figure~\ref{fig:Q3_reset}, HSR and HSR-only still achieve the best performance. This indicates that HSR improves both interaction efficiency and reset efficiency.

\section{Non-Episodic Training Brings Benefits Beyond Reset Efficiency}
\label{sec:benefits}

What are other advantages NSR-MTRL can bring, aside from reducing the labor cost?  Under NSR-MTRL, policies are forced to perform a sequence of tasks non-stop, in a scene that becomes untidier as tasks proceed.  We conjecture that policies fine-tuned with NSR-MTRL are more robust to out-of-domain spatial configurations and long-horizon manipulation tasks than their episodically trained counterparts.  To test this hypothesis, we train both OpenVLA and SpatialVLA with episodic RL ($T=80$) and NSR-MTRL ($T=320$), using 2.9M interaction steps.  We test these models under two settings: (1) independent-task evaluation over out-of-domain configurations, and (2) sequential-task evaluation over out-of-domain configurations.  Results are presented as follows.

\begin{figure}[H]
    \centering
    \includegraphics[width=0.98\textwidth]{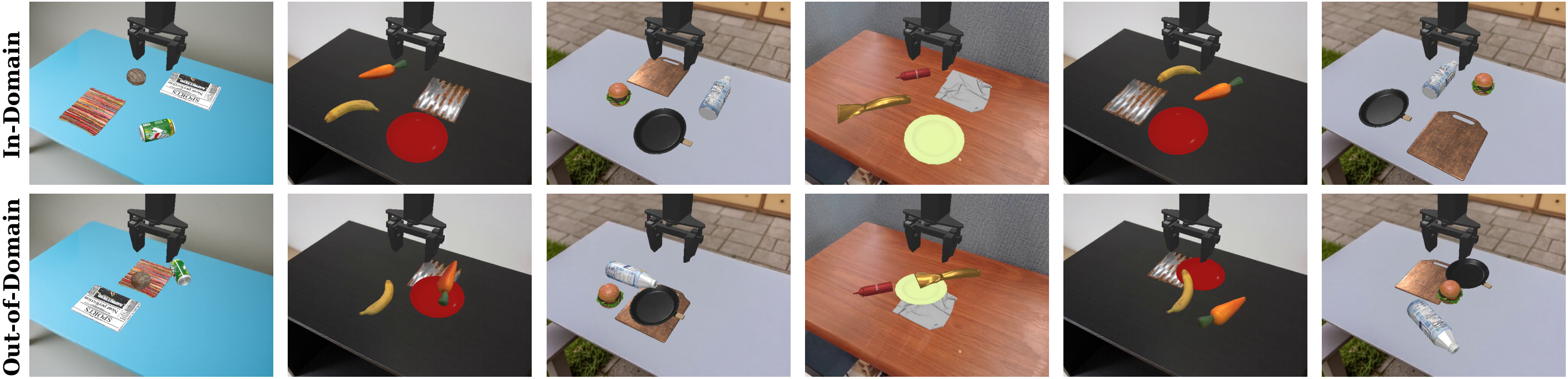}
    \caption{\textbf{In-domain and out-of-domain scene configurations.}
    Out-of-domain scenes include overlapping and cluttered object layouts that do not appear during training.}
    \label{fig:disperse_ood_scene}
\end{figure}

\begin{wraptable}{r}{0.48\linewidth}
    \vspace{-12pt}
    \centering
    \small
    \setlength{\tabcolsep}{6pt}
    \renewcommand{\arraystretch}{1.15}
    \begin{tabular}{l c|cc}
        \toprule
        Model & $T$ & Overlap & Cluttered \\
        \midrule
        \multirow{2}{*}{OpenVLA} & 80  & 0.554 & 0.494 \\
                                 & 320 & \textbf{0.593} & \textbf{0.536} \\
        \midrule
        \multirow{2}{*}{SpatialVLA} & 80  & 0.552 & 0.512 \\
                                    & 320 & \textbf{0.643} & \textbf{0.550} \\
        \bottomrule
    \end{tabular}
    \caption{\textbf{Exposure to dispersed scenes improves spatial generalization.}
    Non-episodic resetting outperforms episodic resetting in both dispersed scenes.}
    \label{tab:disperse_ood}
    \vspace{-10pt}
\end{wraptable}

\subsection{Exposure to Dispersed Scenes Improves Spatial Generalization}
\label{subsec:dispersed}

To evaluate a policy's robustness to out-of-domain initial scene layout, we propose two ways to initialize a scene.  The first is the Overlap setup, placing at least two objects within 12\,cm to each other.  Objects therefore overlap or are in close contact.  The second is the Cluttered setting, where objects are initialized at random positions above the tabletop and fall freely with physics simulation.  This results in natural cluttered scenes.  These two settings are beyond training distribution, which randomly place each object or receptacle on any of 16 positions without overlapping (Figure~\ref{fig:disperse_ood_scene}).

Our evaluation results are summarized in Table~\ref{tab:disperse_ood}.  NSR-MTRL fine-tuning outperforms episodic fine-tuning in both settings. For SpatialVLA, the success rate rises from 55.2\% to 64.3\% in the Overlap setting and from 51.2\% to 55.0\% in the Cluttered setting. For OpenVLA, it rises from 55.4\% to 59.3\% and from 49.4\% to 53.6\%, respectively.  Notably, while episodic training ($T=80$) use four times as many scene resets as NSR-MTRL ($T=320$), it is less robust to out-of-domain configurations than non-episodic counterpart.  Our results demonstrate that exposing policies to more diverse configurations enhances their out-of-domain generalization, and NSR-MTRL naturally enables this capability.

\subsection{Continual Training Improves Sequential Task Execution}
\label{subsec:sequential}

We evaluate whether NSR-MTRL enhances a policy's robustness to both long-horizon robotic manipulation and out-of-domain configurations.  In the following experiments, policies are tasked to perform four tasks in sequence non-stop within scenes, which are initialized under Overlap settings.  For sequential-task evaluation, we consider two cases: one tests on sequences seen during training, and the other tests on unseen sequences.

\begin{figure}[H]
    \centering
    \vspace{-5pt}
    \tb{@{}cc@{}}{0.1}{
    \imw{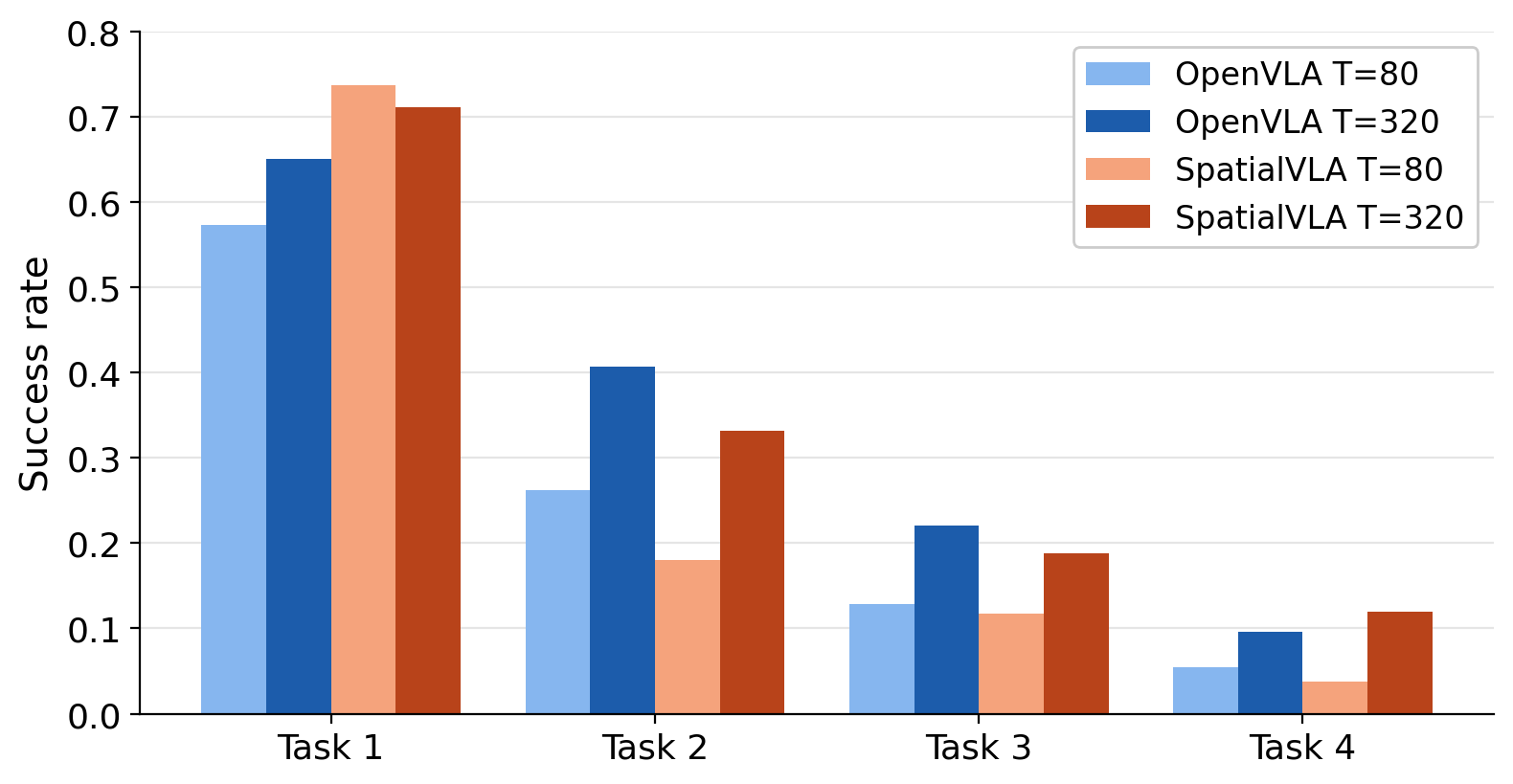}{0.49} &
    \imw{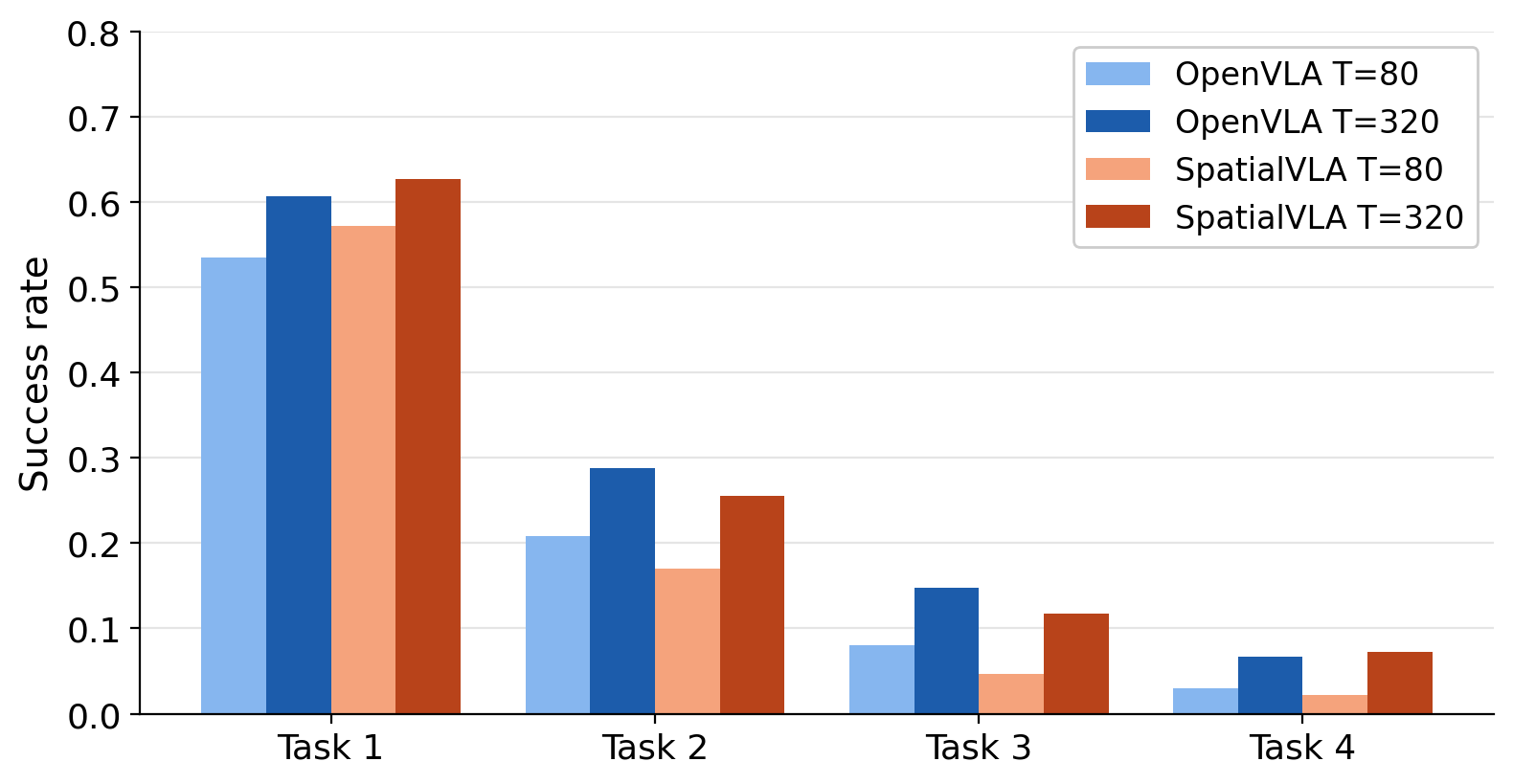}{0.49} \\
    \small Seen Task Sequence &
    \small Unseen Task Sequence \\
    }
    \caption{\textbf{Non-episodic scene reset training improves sequential task execution.} Non-episodic training outperformed episodic training on all subsequent tasks, including both task sequences seen and unseen during training. This improvement was consistent across different VLAs.}
    \label{fig:seq_eval}
    \vspace{-5pt}
\end{figure}

Our results are summarized in Figure~\ref{fig:seq_eval}.  While policies fine-tuned with episodic RL and NSR-MTRL achieve comparable performance on the first task, their performance gap enlarges with the following tasks.  NSR-MTRL fine-tuned policies have an obvious lead over episodically fine-tuned counterparts.  On seen task sequences, OpenVLA reaches 40\% on the second task with $T=320$, compared to 26\% with $T=80$.  SpatialVLA reaches 33\% with $T=320$, compared to 18\% with $T=80$. After proceeding to the fourth task, NSR-MTRL achieves larger performance gain: it nearly doubles and triples the success rate of OpenVLA and SpatialVLA, both fine-tuned with $T=80$.  We observe the similar trend on unseen sequences.

\section{Related Work}
\label{related work}

\noindent\textbf{Fine-tuning VLA with RL.}
Vision-Language Action (VLA) models~\citep{intelligence2025pi, kim2024openvla, o2024open, ankile2025residual, octo2024, pi05_2025, black2024pi0} learn general robot policies across tasks using pre-trained vision-language foundation models. Despite their strong representational capacity, they share a key limitation of behavior cloning: poor robustness to out-of-distribution scenarios. Recent work shows that reinforcement learning fine-tuning can significantly improve generalization in perception, semantics, and execution~\citep{liu2025can}. Several approaches target long-horizon generalization, including task decomposition with dense rewards, action chunking with value estimation, world-model-based verification, and exploration-driven strategies~\citep{zhang2025reinbot, huang2025corft, li2025vlarft, lu2025vla}. However, existing methods assume episodic training, while our benchmark studies VLA fine-tuning in a non-episodic setting.

\noindent\textbf{Non-episodic RL.}
Real-world RL is limited by the need for human intervention to reset environments, and prior benchmarks demonstrate that episodic algorithms do not directly extend to non-episodic settings~\citep{liu2023libero, sharma2022autonomous}. Existing approaches address this limitation in several ways. Model-based methods improve exploration without resets~\citep{yang2025reset, sukhija2024neorl}. Other work jointly learns task and reset policies, explicitly handling irreversible states~\citep{lee2024curriculum, patil2024switching, sharma2023self, kim2023demofree, walke2022scratch, sharma2022state, sharma2022autonomous, zhang2023reset, xie2022ask, eysenbach2018leave}. Another line of work formulates reset-free learning as multi-task learning, where auxiliary tasks implicitly induce resets~\citep{gupta2022demonstration, xu2022dexterous, gupta2021reset, ha2020walk}. Hybrid approaches have also been explored~\citep{yang2024robot}. In contrast, our benchmark quantifies reset efficiency through the number of resets and highlights the core challenges of reset-free learning.

\section{Conclusion}

We introduce \benchmark{}, a simulation benchmark for RL fine-tuning of VLAs under non-episodic scene resets. Using \benchmark{}, we find that naive non-episodic fine-tuning fails because object configurations disperse and the end-effector drifts from its pre-training pose. Targeted resets recover much of this loss. Resetting the end-effector at each task switch removes the pre-training bias at no human cost, scheduling scene resets densely early in training eases learning, and resetting only unrecoverable scenes restores learning under long reset intervals. Beyond reset efficiency, non-episodic training yields policies that generalize better to dispersed scenes and perform better on task sequences.

\paragraph{Limitations.} 
Heuristic scene reset relies on privileged simulator state to detect unrecoverable scenes, and replacing it with a perception-based detector remains future work. Scene drift also continues to limit learning under long reset intervals. Addressing this drift and extending \benchmark{} to contact-rich tasks are important directions for future work.

\section*{Acknowledgment}
This research was supported by the National Science and Technology Council (NSTC), Taiwan, under Grants 114-2222-E-002-008-, 115-2222-E-002-006-, 115-2634-F-002-001-,  115-2218-E-002-026-, and 115-2634-F-002-012-.

\bibliographystyle{iclr2027_conference}
\bibliography{bibs/main}

\appendix
\newpage

\section{NSR-MTRL Algorithm}
\label{sec:appendix_train_algorithm}
Our full algorithm, summarized in Algorithm~1, illustrates the complete training procedure of NSR-MTRL.
Training progresses over a continuous interaction stream and is structured around task-switch intervals.
At each task switch, the agent may trigger one or more reset mechanisms according to the specified reset strategies.
Environment resets are not tied to episode terminations, but are instead applied selectively.
\begin{algorithm}[H]
\small
\caption{Non-episodic Scene Reset Multi-Task RL (NSR-MTRL)}
\label{alg:rfmtrl}
\resizebox{0.95\linewidth}{!}{%
\begin{minipage}{\linewidth}
\begin{algorithmic}[1]

\STATE \textbf{Input:} task set $\mathcal{T}$, policy $\pi_\theta$, value $V_\phi$
\STATE \textbf{Input:} total interaction steps $S$, task switch interval $K$, update interval $U$
\STATE \textbf{Input:} scene reset interval $T$, reset budget $B$
\STATE \textbf{Optional:} end-effector reset (EER), unrecoverable-state detector $\mathbb{I}_{\text{bad}}(s)$ (HSR)

\STATE Initialize environment state $s_0$, step counter $n \leftarrow 0$
\STATE Initialize reset counter $b \leftarrow 0$, buffer $\mathcal{D} \leftarrow \emptyset$

\WHILE{$n < S$ \textbf{and} $b < B$}

    \STATE \textbf{Task switch:} sample instruction $\tau \sim \textsc{TaskScheduler}(\mathcal{T})$

    \FOR{$t = 1$ to $K$}
        \STATE Observe $o_t$ and instruction $\tau$
        \STATE Sample action $a_t \sim \pi_\theta(a \mid o_t, \tau)$
        \STATE Step environment $s \leftarrow f(s, a_t)$ and receive reward $r_t$
        \STATE Store $(o_t, \tau, a_t, r_t)$ into buffer $\mathcal{D}$
        \STATE $n \leftarrow n + 1$
        \IF{$(n \bmod U) = 0$}
            \STATE Compute returns and advantages using $V_\phi$
            \STATE Update policy $\pi_\theta$ and value $V_\phi$ on $\mathcal{D}$
            \STATE Clear buffer $\mathcal{D} \leftarrow \emptyset$
        \ENDIF
    \ENDFOR

    \IF{EER enabled}
        \STATE Reset the end-effector to its canonical pose
    \ENDIF

    \IF{HSR enabled \textbf{and} $\mathbb{I}_{\text{bad}}(s) = 1$}
        \STATE Apply scene reset; $b \leftarrow b + 1$
    \ELSIF{$(n \bmod T) = 0$}
        \STATE Apply scene reset; $b \leftarrow b + 1$
    \ENDIF

\ENDWHILE

\end{algorithmic}
\end{minipage}
}
\end{algorithm}

\newpage

\section{Hyperparameter details}
\label{sec:appendix_train_details}
We detail the training configuration and experiment setup used in our experiments.
Table~\ref{tab:training_hyperparameters} reports the hyperparameters used across all experiments,
covering PPO optimization, the vision-language-action (VLA) model configuration based on OpenVLA and SpatialVLA,
and additional training settings related to reset-free multi-task learning.
\begin{table}[H]
\centering
\begin{tabular}{l c}
\toprule
\textbf{Hyperparameter} & \textbf{Value} \\
\midrule

\multicolumn{2}{l}{\textbf{PPO}} \\
\midrule
PPO epochs per update & 1 \\
Clip ratio ($\epsilon$) & 0.2 \\
Discount factor ($\gamma$) & 0.99 \\
GAE parameter ($\lambda$) & 0.95 \\
Entropy coefficient & 0.0 \\
Gradient accumulation steps & 20 \\

\midrule
\multicolumn{2}{l}{\textbf{Vision-language-Action (VLA) model}} \\
\midrule
Backbone & OpenVLA-7B, SpatialVLA-4B \\
LoRA rank & 32 \\
Optimizer & AdamW \\
Learning rate (VLA backbone) & $1 \times 10^{-4}$ \\
Learning rate (value head) & $3 \times 10^{-3}$ \\
Adam $\beta_1$, $\beta_2$ & 0.9, 0.999 \\
Action sampling temperature (train) & 1.0 \\
Action sampling temperature (eval) & 0.6 \\

\midrule
\multicolumn{2}{l}{\textbf{Other Training Settings}} \\
\midrule
Number of parallel environments & 64 \\
Training horizon (steps) & 320 \\
Instruction switch interval & 80 \\
Training update interval & 160 \\

\bottomrule
\end{tabular}
\vspace{1em}

\caption{Training hyperparameters used in all experiments.}
\label{tab:training_hyperparameters}
\end{table}

\newpage

\section{Reset Strategies}
\label{app:reset_methods}

\begin{figure*}[ht!]
    \centering
    \includegraphics[width=\textwidth]{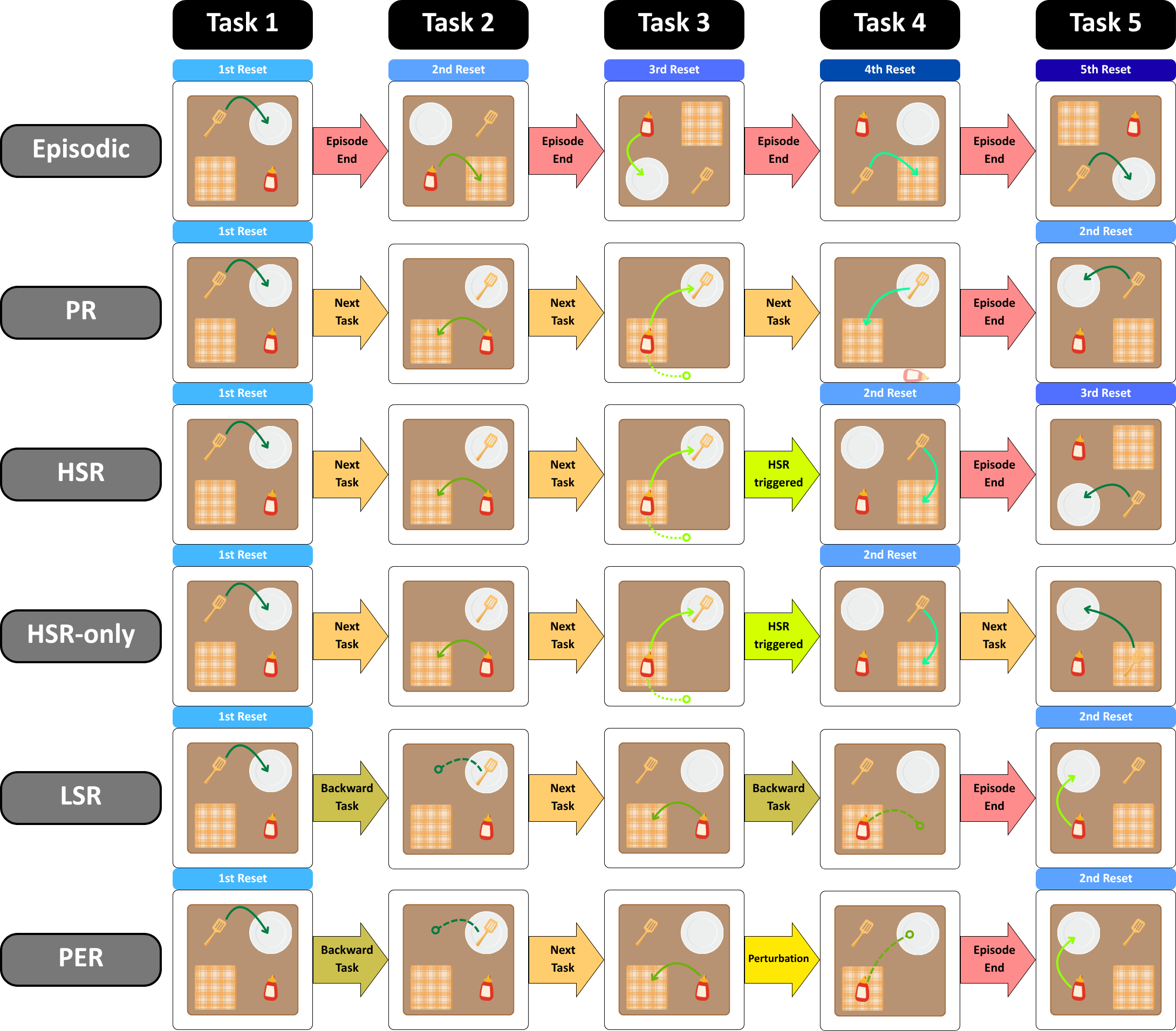}
    \caption{ \textbf{Visualization of reset-free methods.} We compare five methods: (1) periodic reset (PR), which resets the scene every $T=2560$ steps; (2) heuristic scene reset (HSR), which detects unrecoverable states before each task switch and resets only when necessary; (3) heuristic scene reset without periodic reset (HSR-only), which uses heuristic resets without periodic resets, avoiding unnecessary resets in recoverable states; (4) learned scene reset (LSR), which learns reset behavior as an auxiliary task within a single policy, eliminating external reset logic at the cost of interaction budget; and (5) perturbation policy (PER), which extends LSR with a second auxiliary task that perturbs the scene, with the two tasks sampled uniformly at random.}
    \label{fig:Q4_baseline_methods}
\end{figure*}

Figure~\ref{fig:Q4_baseline_methods} illustrates the scene reset strategies compared in Section~\ref{experiments}. Each row shows a sequence of five tasks executed in the same scene. A blue label above a task marks a scene reset, and the arrow between two tasks indicates how the policy proceeds to the next one. All non-episodic strategies use EER, so the end-effector returns to its canonical pose at every task transition.

\paragraph{Periodic Reset (PR).}
The scene is reset every $T=2560$ steps, and the policy proceeds directly to the next task between resets. Objects remain wherever the previous task left them, even when they have left the reachable workspace or fallen off the table.

\paragraph{Heuristic Scene Reset (HSR).}
Before each task switch, HSR checks whether the scene has become unrecoverable. We consider a scene unrecoverable when an object has fallen off the table or lies outside the reachable workspace. If so, HSR triggers a scene reset before the next task. HSR also keeps the periodic reset every $T=2560$ steps. HSR-only uses the same check but removes the periodic reset, so the scene is reset only when it becomes unrecoverable.

\paragraph{Learned Scene Reset (LSR).}
Following \citet{eysenbach2018leave}, LSR adds an auxiliary reset task, such as ``put the [object] on the table''. The policy alternates between a forward task and this backward task, so that it learns to return objects toward task-relevant configurations. This removes external reset logic but spends part of the interaction budget on reset behavior. LSR keeps the periodic reset every $T=2560$ steps.

\paragraph{Perturbation Policy (PER).}
Following \citet{zhu2020ingredients}, PER extends LSR with a second auxiliary task that perturbs the scene, such as ``put the [object] on the other plate''. The reset task and the perturbation task are sampled uniformly at random. PER aims to expose the policy to more diverse object configurations. It also keeps the periodic reset every $T=2560$ steps.

\end{document}